\UseRawInputEncoding
\documentclass{article}

\usepackage[utf8]{inputenc}
\usepackage[margin=2.5cm]{geometry}
\usepackage{amsmath}
\usepackage{graphicx}
\usepackage{hyperref}
\usepackage{authblk}
\usepackage{setspace}
\usepackage{natbib}
\usepackage{booktabs}
\usepackage{caption}

\hypersetup{
    colorlinks=true,
    linkcolor=blue,
    filecolor=magenta,      
    urlcolor=cyan,
    citecolor=blue,
}
\begin{document}

\title{\textbf{A machine learning framework integrating seed traits and plasma parameters for predicting germination uplift in crops}}

\author{Saklain Niam}
\author{Tashfiqur Rahman}
\author{Md. Amjad Patwary}
\author{Mukarram Hossain}

\affil[1]{Department of Food Engineering and Tea Technology, Shahjalal University of Science and Technology, Sylhet 3114, Bangladesh}

\date{}

\maketitle

\noindent\textbf{Contact}\\
Saklain Niam: 
\href{mailto:saklain35@student.sust.edu}{saklain35@student.sust.edu},
Md. Amjad Patwary: \href{mailto:amjad-fet@sust.edu}{amjad-fet@sust.edu},

\begin{abstract}
Cold plasma (CP) is an eco-friendly method to enhance seed germination, yet outcomes remain difficult to predict due to complex seed--plasma--environment interactions. This study introduces the first machine learning framework to forecast germination uplift in soybean, barley, sunflower, radish, and tomato under dielectric barrier discharge (DBD) plasma. Among the models tested (GB, XGB, ET, and hybrids), Extra Trees (ET) performed best (R\textsuperscript{2} = 0.919; RMSE = 3.21; MAE = 2.62), improving to R\textsuperscript{2} = 0.925 after feature reduction. Engineering analysis revealed a hormetic response: negligible effects at $<$7 kV or $<$200 s, maximum germination at 7--15 kV for 200--500 s, and reduced germination beyond 20 kV or prolonged exposures. Discharge power was also a dominant factor, with germination rate maximizing at $\geq$100 W with low exposure time. Species and cultivar-level predictions showed radish (MAE = 1.46) and soybean (MAE = 2.05) were modeled with high consistency, while sunflower remained slightly higher variable (MAE = 3.80). Among cultivars, Williams (MAE = 1.23) and Sari (1.33) were well predicted, while Arian (2.86) and Ny\'{\i}rs\'{e}gi fekete (3.74) were comparatively poorly captured. This framework was also embedded into MLflow, providing a decision-support tool for optimizing CP seed germination in precision agriculture.
\end{abstract}


\section*{Abbreviations}
\begin{tabular}{ll}
CP & Cold Plasma\\
DBD & Dielectric Barrier Discharge\\
RONS & Reactive Oxygen and Nitrogen Species\\
ET & Extra Trees\\
GB & Gradient Boosting\\
XGB & Extreme Gradient Boosting\\
HM & Hybrid Model\\
CV & Cross-Validation\\
LOCO & Leave-One-Cultivar-Out (validation)\\
LOGO & Leave-One-Genotype-Out (validation, synonym for LOCO in text)\\
PI & Permutation Importance\\
SHAP & SHapley Additive exPlanations\\
ML & Machine Learning\\
AI & Artificial Intelligence\\
MLflow & Machine Learning Lifecycle Platform (experiment tracking and deployment)\\
R² & Coefficient of Determination\\
RMSE & Root Mean Square Error\\
MAE & Mean Absolute Error\\
MAPE & Mean Absolute Percentage Error\\
GI & Germination Index\\
GP & Germination Potential\\
GR & Germination Rate\\
SOD & Superoxide Dismutase (baseline antioxidant activity)\\
ABA & Abscisic Acid\\
GA & Gibberellins
\end{tabular}

\section{Introduction}

Seed germination is the pivotal first stage of a plant's life cycle, determining whether a viable seed transitions into a healthy seedling under favorable conditions of water, temperature, and oxygen \citep{granata2024, karim2025}. The seed germination rate is a critical indicator of seed quality, as it largely influences the overall yield and economic returns \citep{liu2025, tian2025}. Seeds that fail to germinate or produce weak seedlings not only reduce productivity but also undermine resource efficiency, threatening the resilience of farming systems. With global commitments such as the United Nations' Sustainable Development Goal 2, Zero Hunger by 2030, enhancing seed quality and vigor is a central priority for sustainable agriculture \citep{unicef2021}.

Traditionally, seed priming has relied on chemical, osmotic, or hydropriming methods to improve germination and vigor. While effective in many contexts, these approaches face limitations, including high water use, chemical residues, and inconsistent performance under variable environmental conditions \citep{kakar2023, yang2023}. Against the backdrop of climate instability, soil degradation, and the spread of seedborne pathogens, there is increasing demand for more sustainable, non-chemical seed enhancement technologies. CP treatment has emerged in recent years as one of the most innovative and eco-friendly seed priming approaches \citep{singh2020}. Cold plasma is a partially ionized gas composed of reactive oxygen and nitrogen species (RONS), UV photons, and charged particles, generated at near-ambient temperatures \citep{shahabi2025, waskow2021}. Unlike thermal plasmas, CP operates non-destructively, modifying seed coat properties, enhancing water uptake, and modulating antioxidant defense systems without impairing viability \citep{shahabi2025}. A growing body of experimental research has reported beneficial effects of CP on diverse crops such as soybean, radish, tomato, barley, and sunflower, demonstrating improvements in germination speed, uniformity, and stress tolerance \citep{benabderrahim2024, murali2025}. For instance, soybean, a legume important for global food and oil markets, often exhibits strong germination responses under CP treatment, improving both speed and uniformity \citep{sayahi2024b}. Radish, a fast-growing vegetable, benefits significantly from CP priming, showing improvements in germination rate and seedling development \citep{kanjevac2022}. Tomato, a major horticultural crop, also responds well to CP, with studies indicating enhanced stress tolerance and faster germination under optimal conditions \citep{adhikari2020}. Barley and sunflower, though beneficial in certain conditions, show greater variability in CP responses. Barley's sensitivity to plasma exposure often results in mixed outcomes, while sunflower's larger seed size and thicker coat can sometimes limit treatment efficacy \citep{benabderrahim2024, tamosiune2020}. This variability arises from the complex interplay between seed traits (e.g., vigor, antioxidant capacity), plasma parameters (e.g., voltage, power), and germination environments. Traditional experimental approaches, which vary one or two parameters at a time, are resource-intensive and insufficient for capturing these complexities.

Machine learning (ML) models such as Extra Trees (ET), Gradient Boosting (GB), and Extreme Gradient Boosting (XGB) can effectively capture nonlinear interactions and high-dimensional relationships in seed--plasma--environment data \citep{dissanayake2025, niam2025, wang2025}. However, to date, there has been no research that applies these models to CP priming. Recent ML applications in agriculture, such as predicting crop yields and evaluating seed vigor, highlight its potential as a decision-support tool for sustainable farming \citep{mohan2025, rayhana2025}. Furthermore, integrating ML into experiment-tracking frameworks ensures reproducibility and facilitates the automation of CP priming, transforming it from trial-and-error to data-driven optimization \citep{barreiro2025, pepe2021}.

Although CP priming has shown promise in enhancing seed germination, a significant research gap remains in developing predictive frameworks that can reliably forecast germination outcomes and enable automation through machine learning. Therefore, this study addresses the gap by developing a predictive framework for CP-induced germination uplift using a cross-crop, multi-cultivar dataset of soybean, barley, sunflower, radish, and tomato, and benchmarking ensemble models including GB, XGB, ET, and stacked hybrids. To the best of our knowledge, this is the first machine learning--based predictive framework that integrates biological traits with engineering parameters to forecast seed germination, thereby enabling the integration of cold plasma into precision agriculture and sustainable seed technologies.

\section{Materials and Methods}

\subsection{Experimental data collection}

Datasets were compiled from previously published studies investigating the effect of cold plasma treatment on in-vivo seed germination (supplementary table 1). In total, 196 experimental records were aggregated, covering five crop species with multiple cultivars (table 1). The relatively small dataset reflects the limited number of published studies reporting sufficiently detailed information, as such specific experimental data are often difficult to obtain. Cultivar representation was uneven, with soybean and radish dominating the dataset, while tomato and barley were less represented. This diversity enabled stratified evaluation, allowing models to capture both inter- and intra-species variability in cold plasma sensitivity. The primary plasma treatment source was DBD, although different gas compositions (argon, helium, oxygen, air) and discharge configurations were included. The collected input parameters are as follows:

Seed traits: seed size (mm), seed weight (g), baseline antioxidant activity (SOD), baseline germination rate (\%), germination potential (\%), germination index, and germination days.

Plasma discharge parameters: plasma plate dimensions (length, width, thickness, cm), plasma temperature (°C), seed-to-electrode distance (cm), discharge voltage (kV), frequency (kHz), power (W), pressure (kPa), gas flow rate (L/min), plasma exposure time (s), and discharge gas type: Ar, He, O$_2$, air.

Germination environment: growing temperature (°C) and water per seed (ml / gr).

Output variables: germination uplift (\% relative to untreated control).

\begin{table}[h]
\centering
\caption{Representation of crop species and cultivars included in the dataset.}
\label{tab:table1}
\begin{tabular}{lll}
\toprule
Seed & Cultivar & No. of data \\
\midrule
Barley & Planet & 26 \\
Radish & Raphanus sativus & 36 \\
Soybean & Arian & 20 \\
Soybean & Katoul & 10 \\
Soybean & Saba & 10 \\
Soybean & Sari & 10 \\
Soybean & Williams & 10 \\
Sunflower Seed & Nyírségi fekete & 53 \\
Tomato & Belle F1 & 31 \\
\bottomrule
\end{tabular}
\end{table}

Searches were conducted in Google Scholar and Web of Science, and only in-vivo germination studies were considered. Control data were included wherever reported. When variables were reported in graphical form only, values were digitized manually using PlotDigitizer.com. To avoid data leakage from baseline germination rate to predicted germination rate, the new feature termed germination rate uplift was calculated as:

\begin{equation}
\text{Germination Rate Uplift} = \text{treated seed germination rate} - \text{base germination rate}
\end{equation}

\subsection{Data imputation and outlier detection}

No data points were excluded, as outliers were considered genuine biological variability in CP responses rather than measurement error \citep{matejovic2025, sayahi2024b}. Missing values were imputed using within-species means to maintain biological consistency and prevent data leakage.

\subsection{Data analysis}

The dataset was analyzed to describe both seed traits and cold plasma treatment parameters before model building. Descriptive statistics (mean, standard deviation, minimum, maximum, skewness, and kurtosis) were calculated for all numeric features (supplementary table 2).

Most plasma-related parameters showed wide ranges and positive skewness, reflecting that many experiments were done under moderate conditions, with fewer at very high intensity. Plasma time and frequency had long right tails, while voltage and seed-to-electrode distance were spread more evenly (supplementary figure 2). Some features showed plateau-like behavior: plasma temperature clustered strongly around 25 °C with occasional higher values, and pressure was bimodal, with groups around low-pressure and atmospheric regimes. The target variable, germination rate uplift, had a mean of about 12\% with both positive and negative values. Its distribution was slightly left-skewed, showing that while most treatments led to moderate improvements, a few cases achieved a very low uplift. Negative values confirmed that plasma can sometimes reduce germination.

The scatter plots (figure 1) show linear fits between individual plasma parameters and germination rate uplift. Several parameters displayed modest linear associations with uplift, with R² values ranging mostly between 0.1 and 0.3. For instance, power and frequency showed positive slopes, suggesting that higher power and frequencies were linked to greater uplift in some cases. In contrast, voltage and pressure showed negative slopes, consistent with earlier reports that too much discharge intensity can damage seed coats \citep{mohajer2024a, sayahi2024a}. Gas flow rate showed a no correlations, though prior studies note that gas type and discharge geometry strongly influence reactive species flux to the seed surface \citep{odah2025, torres2025}. These results confirmed that cold plasma effects on germination are not explained by one variable alone, but by combined interactions between seed properties, plasma settings, and growth environment.

\begin{figure}[h]
\centering
\includegraphics[width=0.7\textwidth]{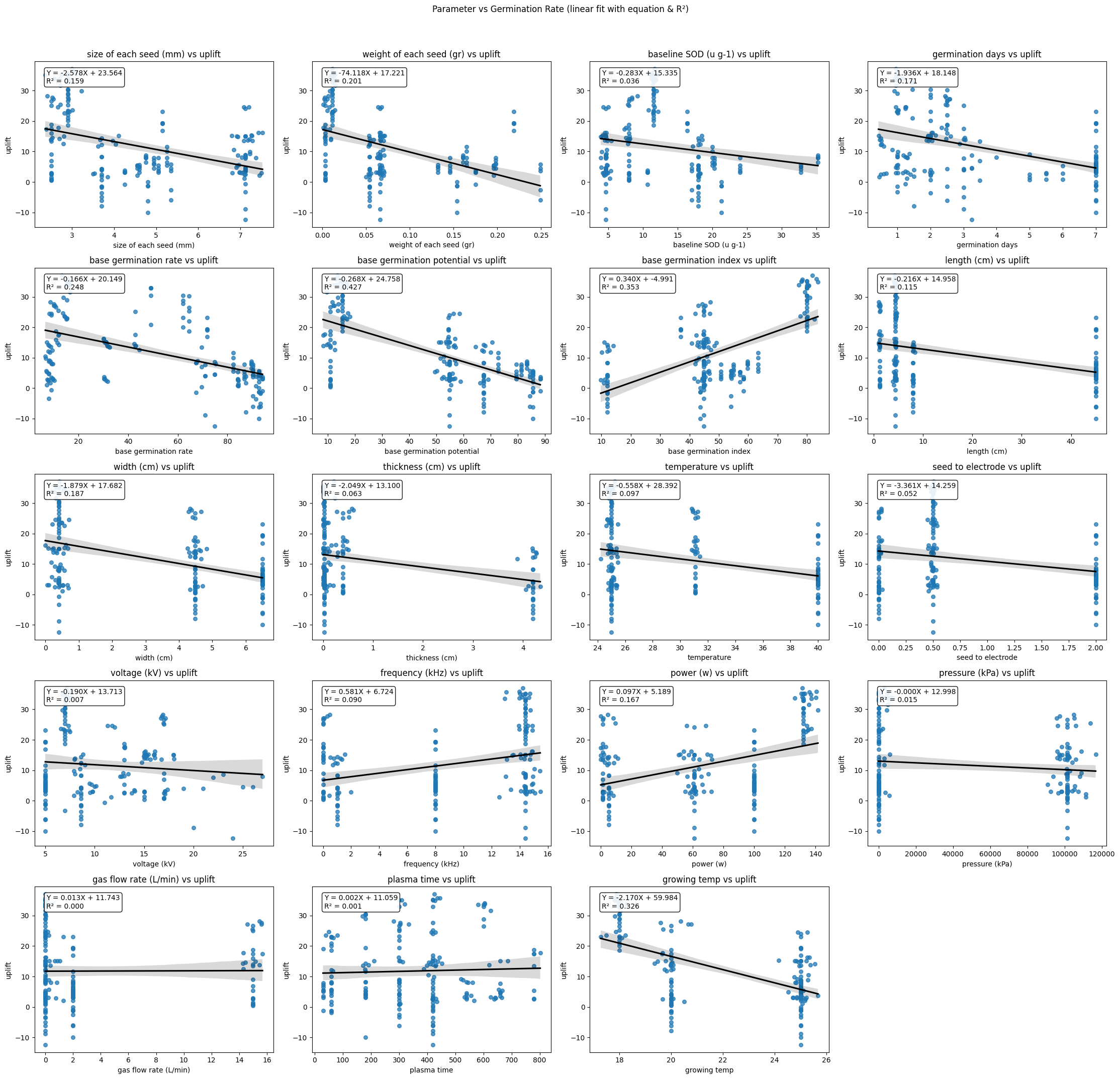}
\caption{Scatter plots of plasma parameters versus germination rate with fitted linear regression lines and coefficients of determination (R²).}
\label{fig:figure1}
\end{figure}

\subsection{Species and cultivar-specific responses to cold plasma priming}

Cold plasma effects on germination varied across species and cultivars. Radish and Belle F1 showed the most consistent positive responses, soybean and tomato moderate but stable gains, while barley and sunflower were highly variable (figure 2a). At the cultivar level (figure 2b), Williams, Sari, and Saba performed reliably, whereas Arian, Nyírségi fekete, Planet, and Katoul were inconsistent, confirming that plasma sensitivity depends on species- and cultivar-specific traits such as seed coat permeability and antioxidant defenses \citep{sayahi2024a, waskow2021}.

\begin{figure}[h]
\centering
\includegraphics[width=\textwidth]{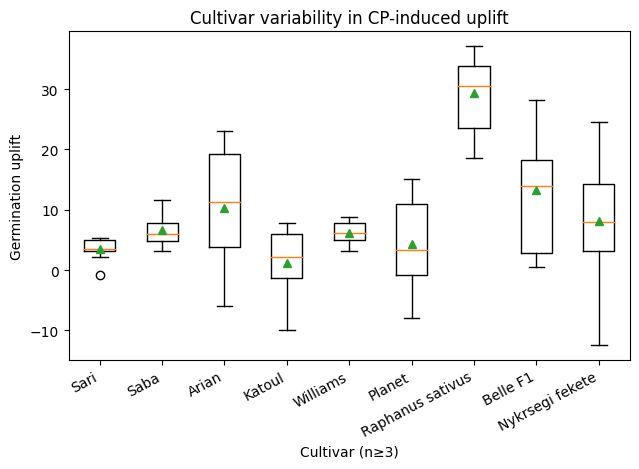}
\caption{Germination rate uplift across a) seed, b) cultivar species following cold plasma treatment.}
\label{fig:figure2}
\end{figure}

\subsection{Proposed workflow}

This study follows a systematic workflow designed to preprocess, optimize, and evaluate predictive models for seed germination estimation (supplementary figure 3).

\subsubsection{Yeo-Johnson transformation}

To mitigate skewness, normalize feature distributions, and reduce the influence of extreme values, the Yeo-Johnson transformation is applied to the input variables. The Shapiro--Wilk test confirms that all measured input variables deviate from normality. After applying Yeo--Johnson transforms, the W statistic increased, indicating improved normality.

\subsubsection{Polynomial feature engineering}

To capture non-linear and interactive effects among cold-plasma priming parameters, a degree-2 polynomial expansion was applied to a small, physics-informed subset of drivers: power (W), plasma time (s), and discharge voltage (kV). For each variable $x_k$, the squared term $x_k^2$ was generated, and for each pair $(x_i, x_j)$, the interaction term $x_ix_j$ was created. The resulting terms include $P^2$, $t^2$, $V^2$, and the pairwise interactions $P \times t$ (delivered energy), $V \times t$ (electric field exposure over time), and $P \times V$ (discharge intensity). This step ensures that higher-order relationships are retained while maintaining interpretability.

\subsubsection{Feature scaling}

All features were standardized with StandardScaler to improve numerical stability and avoid scale dominance. This procedure centers each variable to a zero mean and scales it to unit variance, placing all predictors on a comparable scale and improving the conditioning of optimization algorithms \citep{wongoutong2024}. Equation (2) was applied to standardize the variables.

\begin{equation}
k' = \frac{k - \bar{k}}{\sigma}
\end{equation}

Where, $k'$ = standardized data, $k$ = mean value, $\bar{k}$ = each value, $\sigma$ = standard deviation of the data.

\subsubsection{Train-test split and random state}

To evaluate the robustness of the models, the dataset was divided into training and testing sets in a 70:30 proportion \citep{arnob2025}. A fixed random state of 3 was applied to maintain consistency and reproducibility of results across different experimental runs.

\subsubsection{Hyperparameter optimization and model training}

Machine learning models, including ET, GB, XGB, and hybrid ensembles, were trained due to their effectiveness. Hyperparameter optimization was performed using GridSearchCV with five-fold cross-validation to optimize key parameters. Model-specific hyperparameters, including learning rate and the base estimators for stacking models, were fine-tuned. The root mean squared error (RMSE) serves as the primary optimization criterion to ensure model robustness.

\subsubsection{Cross-validation}

Cross-validation was used to ensure reliable model evaluation and avoid overfitting \citep{tougui2021}. A 5-Fold CV on 70\% training data allowed each sample to be used for both training and validation, with averaged results providing robust performance estimates \citep{gorriz2024}. Final testing was performed on the 30\% unseen data to assess generalization. In addition, a Leave-One-Cultivar-Out (LOCO) strategy was applied, where all samples from one cultivar were reserved for testing, simulating deployment to entirely new genetic backgrounds.

\subsubsection{Evaluation metrics}

To evaluate the performance of machine learning models for seed germination, this study employed 3 key statistical metrics: Mean Absolute Error (MAE), Root Mean Square Error (RMSE), and Coefficient of Determination (R²). The mathematical representations of these indicators are detailed in supplementary table 3.

\subsubsection{Model management and workflow}

To ensure reproducibility and deployment readiness, the best model's predictive framework was implemented as an end-to-end MLflow pipeline. This pipeline integrates data ingestion, preprocessing, model training, and evaluation of the error metrics.

MLflow automatically tracked hyperparameters, performance metrics, feature importance values, and trained model artifacts across all experimental runs. This enabled transparent comparison of models, traceability of results, and prevention of undocumented changes during optimization. Supplementary figure 4 presents the workflow, highlighting the sequential flow from data ingestion to deployment, with MLflow serving as an experiment-tracking layer.

\subsection{Germination rate prediction using ET, XGB, and GB}

Figure 3 illustrates the workflow for model training. The dataset was used to train ET, XGB, and GB models with systematic hyperparameter tuning, and the optimal values are summarized in table 2. Model performance was evaluated using standard metrics, after which the best-performing model was preserved and deployed to predict germination uplift under unseen cold plasma treatments.

\begin{figure}[htbp] 
    \centering
    \includegraphics[width=\textwidth]{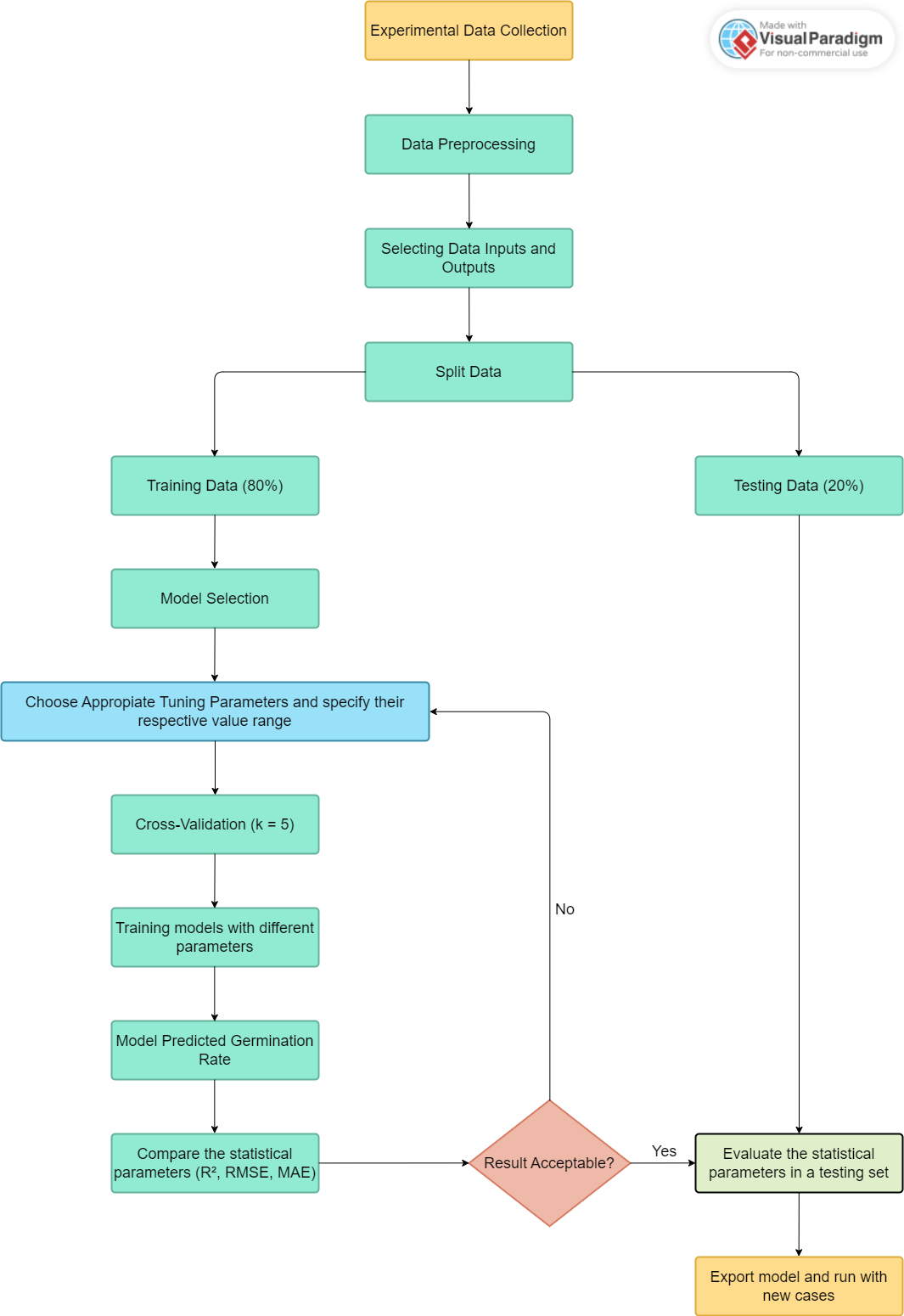}
    \caption{Machine learning workflow.}
    \label{fig:figure3}
\end{figure}

\begin{table}[h]
\centering
\caption{Hyperparameter values of ET, GB, and XGB.}
\label{tab:table2}
\begin{tabular}{ll}
\toprule
Model & Best parameters \\
\midrule
Extra Trees & n\_estimator: 400 \\
 & max\_depth: None \\
 & min\_sample\_leaf: 2 \\
 & min\_sample\_split: 4 \\
\midrule
Gradient Boosting & n\_estimator: 500 \\
 & learning\_rate: 0.01 \\
 & min\_sample\_split: 2 \\
 & min\_sample\_leaf: 6 \\
 & max\_depth: None \\
\midrule
XGBoost & n\_estimator: 300 \\
 & learning\_rate: 0.05 \\
 & max\_depth: None \\
 & sub\_sample: 0.4 \\
 & col\_sample\_bytree: 1 \\
 & min\_child\_weight: 2 \\
 & gamma: 5 \\
\bottomrule
\end{tabular}
\end{table}

\subsection{Germination rate prediction using hybrid models}

To reduce variance and exploit complementary model biases, four stacking ensemble was built using the top three models (ET, GB, XGB; Supplementary figure 5). The hybrid configurations tested were: HM1 (ET+GB), HM2 (ET+XGB), HM3 (GB+XGB), and HM4 (ET+GB+XGB). Base learners were tuned via cross-validation, and RidgeCV ($\alpha \in \{0.1, 1, 10, 100\}$) served as the meta-learner. Training used a 5-fold out-of-fold strategy to generate level-2 inputs, with final predictions made on the held-out test set. This ensured strict data separation, minimized information leakage, and improved generalization across plasma treatments.

\subsection{SHAP}

To interpret model predictions, we applied the SHAP (SHapley Additive exPlanations) framework, which assigns each feature a contribution value toward the predicted germination uplift. For tree-based models like Extra Trees, SHAP efficiently approximates Shapley values, enabling both global and local feature attribution. In this study, SHAP was applied to the best-performing model on the test set, where global values highlighted the most influential predictors and local values revealed cultivar-specific variations. This provided a transparent link between seed traits, plasma parameters, and model outputs, distinguishing biological drivers (seed vigor) from engineering levers (voltage, power, time).

\section{Results and discussion}

\subsection{Comparison of baseline models}

Table 3 shows the performance of the three baseline regressors (ET, GB, XGB) in predicting germination uplift. All models fit the training data well (R² = 0.986--0.990). On the test set, ET performed best (RMSE = 3.23, MAE = 2.63, R² = 0.919), whereas GB showed moderate accuracy (RMSE = 4.20, MAE = 3.18, R² = 0.863) with a larger train-test gap. XGB, though it achieved the highest in-sample R², had the weakest generalization, indicating overfitting. With mean uplift at 11.8\% (SD 11.4\%), ET's MAE $\approx$2.6 ($\sim$0.23 SD) indicates strong explanatory power.

\begin{table}[h]
\centering
\caption{Comparison of ET, XGB, and GB models.}
\label{tab:table3}
\begin{tabular}{llccc}
\toprule
Model & Split & RMSE & MAE & R² \\
\midrule
ET & Train & 1.34 & 0.85 & 0.986 \\
 & Test & 3.21 & 2.62 & 0.920 \\
\midrule
XGB & Train & 1.37 & 0.99 & 0.985 \\
 & Test & 4.87 & 3.85 & 0.817 \\
\midrule
GB & Train & 1.26 & 0.86 & 0.988 \\
 & Test & 4.12 & 3.18 & 0.869 \\
\bottomrule
\end{tabular}
\end{table}

Figures 4a-c illustrates that the predicted uplift (red line) closely follows the ground-truth trajectory (black line). Discrepancies are largest in the low-to-mid uplift range (--10\% to $\sim$10\%), where oscillations indicate local variability. Predictions stabilize in the mid-to-high uplift range ($>$15\%), with ET closely tracking the ground-truth curve. XGB shows more misalignment in the low-to-mid range, while GB exhibits larger fluctuations at higher uplift values. Overall, ET provides the best balance of fit and generalization, while GB and XGB serve as complementary alternatives in ensemble models \citep{ghazwani2023, ser2023}.

\begin{figure}[h]
\centering
\includegraphics[width=0.7\textwidth]{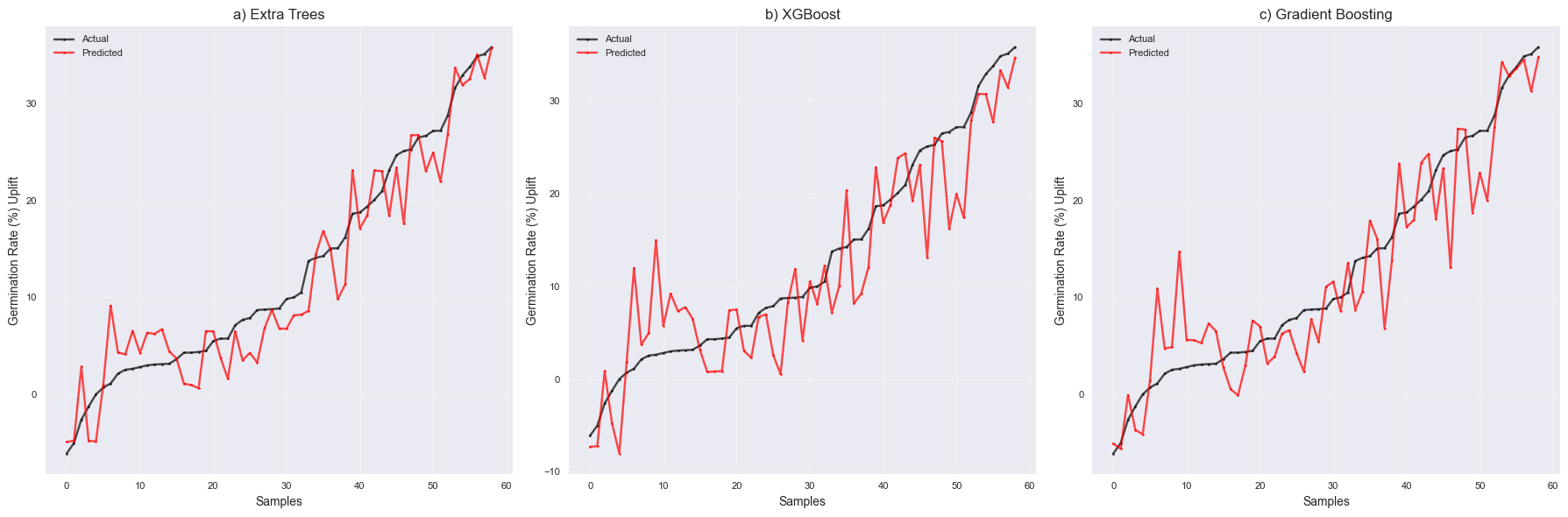}

\vspace{1em} 
\includegraphics[width=0.7\textwidth]{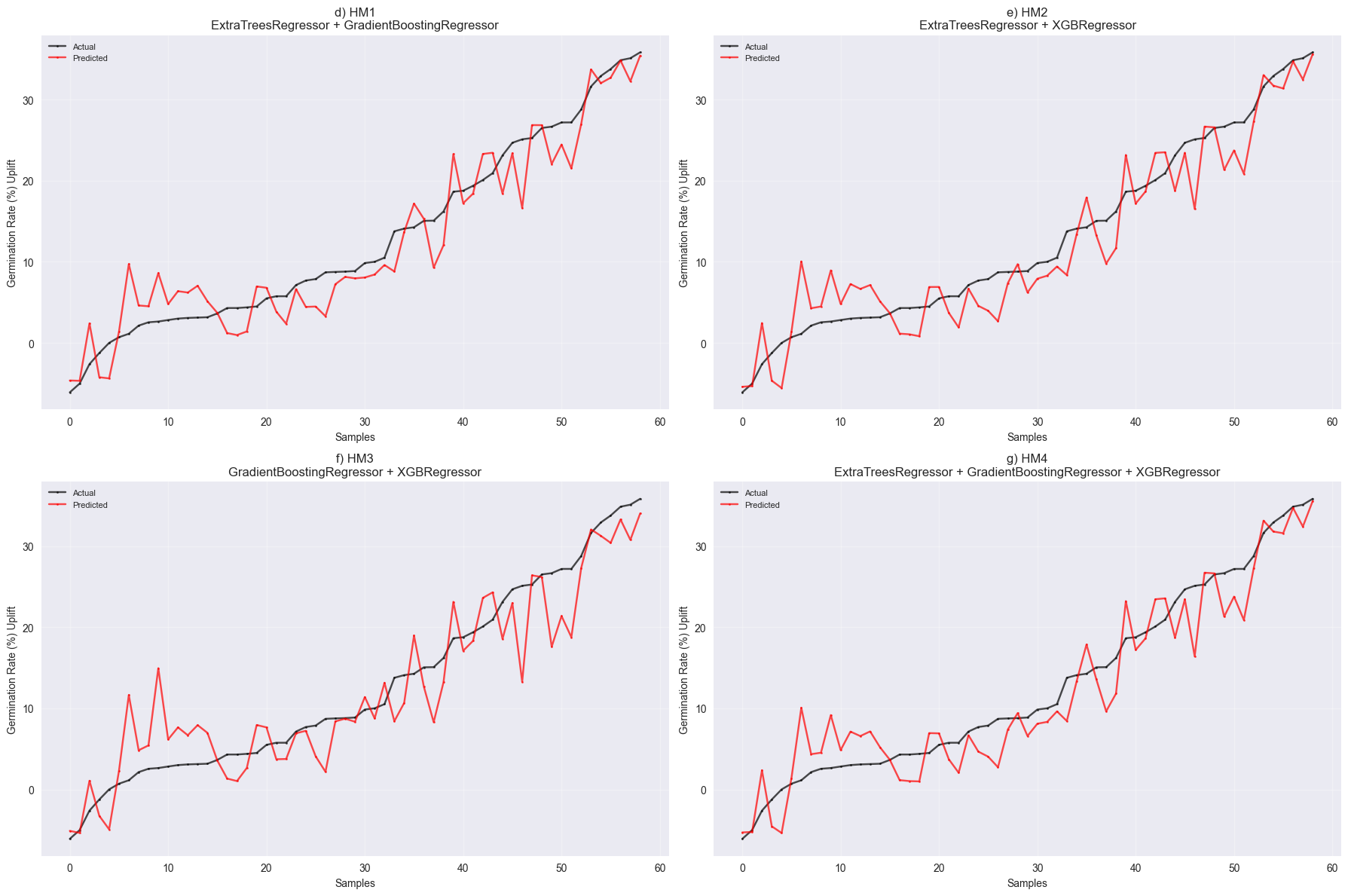}

\caption{Actual vs. Predicted among the ET, XGB, GB and hybrid models (HM1, HM2, HM3, and HM4) models.}
\label{fig:figure4_combined}
\end{figure}

\subsection{Comparison of hybrid models}

The four stacking hybrids (HM1, HM2, HM3, and HM4) were built on single models and performed well on both training and test data (Table 4). HM1 achieved the best balance (RMSE = 3.28, MAE = 2.65, R² = 0.917), slightly lower than ET. HM2 also showed strong accuracy (RMSE = 3.79, MAE = 3.00, R² = 0.889), while HM4 remained stable but weaker (RMSE = 3.95, MAE = 3.14, R² = 0.879). HM3 underperformed, indicating that boosting-only hybrids lack the complementary diversity seen in those including ET.

\begin{table}[h]
\centering
\caption{Comparison of HM1, HM2, HM3, and HM4 models.}
\label{tab:table4}
\begin{tabular}{llccc}
\toprule
Model & Split & RMSE & MAE & R² \\
\midrule
HM1 & Train & 1.32 & 0.85 & 0.986 \\
 & Test & 3.28 & 2.65 & 0.917 \\
\midrule
HM2 & Train & 1.26 & 0.83 & 0.988 \\
 & Test & 3.79 & 3.00 & 0.889 \\
\midrule
HM3 & Train & 1.39 & 0.99 & 0.985 \\
 & Test & 4.84 & 3.68 & 0.818 \\
\midrule
HM4 & Train & 1.31 & 0.84 & 0.987 \\
 & Test & 3.95 & 3.14 & 0.879 \\
\bottomrule
\end{tabular}
\end{table}

The sorted-by-actual line plots (figure 4d-g) support these findings, showing that predicted uplift curves in all four hybrids follow the ground-truth trajectory. The largest deviations occur in the low-to-mid uplift range (--10\% to $\sim$10\%), with sharper oscillations and larger residuals. In the mid-to-high uplift range ($>$20\%), predictions align more consistently with the ground-truth. Overall, ET-inclusive hybrids (HM1, HM2, HM4) show smoother tracking and fewer extreme deviations, consistent with their lower error scores. In contrast, HM3 exhibits pronounced fluctuations and overshooting in the low uplift range, reflecting its weaker performance.

\subsection{Ranking of models}

Table 5 compares the single models and the hybrids using test-set R² (higher is better) and error metrics RMSE/MAE (lower is better). Models are ranked by descending R² and ascending errors; ties are resolved by RMSE and then MAE.

\begin{table}[h]
\centering
\caption{Ranking of models.}
\label{tab:table5}
\begin{tabular}{lccccccc}
\toprule
Model & R² & RMSE & MAE & R² Rank & RMSE Rank & MAE Rank & Overall Rank \\
\midrule
ET & 0.920 & 3.21 & 2.62 & 1 & 1 & 1 & 1 \\
HM1 & 0.917 & 3.28 & 2.65 & 2 & 2 & 2 & 2 \\
HM2 & 0.889 & 3.79 & 3.00 & 3 & 3 & 3 & 3 \\
HM4 & 0.879 & 3.95 & 3.14 & 4 & 4 & 4 & 4 \\
GB & 0.869 & 4.12 & 3.18 & 5 & 5 & 5 & 5 \\
HM3 & 0.818 & 4.84 & 3.68 & 6 & 6 & 6 & 6 \\
XGB & 0.817 & 4.87 & 3.85 & 7 & 7 & 7 & 7 \\
\bottomrule
\end{tabular}
\end{table}

Overall, ET secures the top position with the highest R² and the lowest errors. Close behind in second position, HM1 achieves nearly identical scores, confirming that stacking ET with GB provides virtually no loss in accuracy. HM2 (rank = 3) and HM4 (rank = 4) follow, both maintaining R² values close to 0.88--0.89 with modestly higher errors. Among the single boosting models, GB ranks 5th, while XGB performs the weakest. The hybrid of boosting-only methods (HM3) also underperforms relative to ET-based ensembles, with R² dropping to 0.818 and error levels close to XGB. Taken together, these results highlight that ET and ET-inclusive hybrids (HM1--HM4) dominate the top rankings, while boosting-only approaches (GB, XGB, HM3) consistently lag behind, consistent with prior studies showing that ET often outperform traditional boosting methods in regression tasks due to their enhanced randomness and reduced overfitting tendencies \citep{ghazwani2023, mohammed2023}.

\subsection{Evaluation of the proposed methodology without standardization and polynomial feature expansion}

To benchmark against the earlier scaled results, models were also evaluated on the raw feature space without any standardization or polynomial expansion (supplementary figure 6). According to figure 5, in this unscaled setting, the ET regressor again achieved the strongest performance among base models, but with substantially weaker accuracy (RMSE = 6.91, MAE = 5.02, R² = 0.63). GB performed slightly worse (RMSE = 7.44, MAE = 5.13, R² = 0.57), while XGB showed the poorest fit (RMSE = 8.60, MAE = 5.87, R² = 0.43). Hybrid ensembles offered only marginal gains, with the best variant (HM1) reaching RMSE = 6.96, MAE = 5.03, and R² = 0.62, comparable to ET but not surpassing it.

Compared with the scaled models presented earlier, these results confirm that preprocessing was necessary \citep{visser2024}. Scaling and polynomial feature expansion reduced errors by more than half and increased R² above 0.90, whereas unscaled models explained only 40--60\% of the variance in germination uplift. This contrast highlights that while tree-based ensembles can extract some structure from raw features, their full predictive power is only realized once input distributions are normalized and nonlinear interactions are incorporated.

\begin{figure}[h]
\centering
\includegraphics[width=0.9\textwidth]{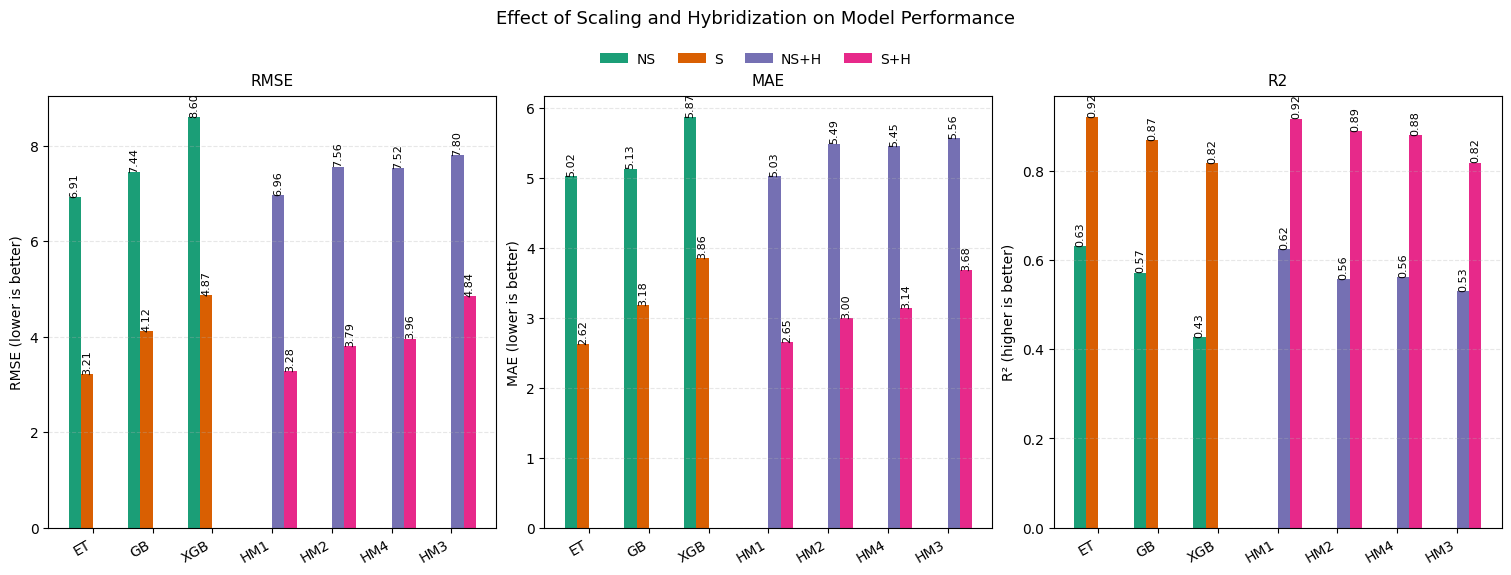}
\caption{Comparison of model performance with and without feature scaling and hybrid stacking. Bars show test set RMSE, MAE, and R² for Extra Trees (ET), Gradient Boosting (GB), XGBoost (XGB), and hybrid models (HM1--HM4) under four settings: no scaling (NS), scaling (S), no scaling with hybridization (NS+H), and scaling with hybridization (S+H).}
\label{fig:figure5}
\end{figure}

\subsection{Feature importance and rebuilding ET model on top features}

We calculated permutation importance (PI) by permuting each feature in the validation set and measuring the resulting drop in model performance, averaging over five repeats (supplementary figure 7) \citep{khan2025}. The PI for ET is shown in Figure 6, where the model was subsequently retrained using the smallest subset of features whose cumulative PI reached 95\%.

\begin{figure}[h]
\centering
\includegraphics[width=\textwidth]{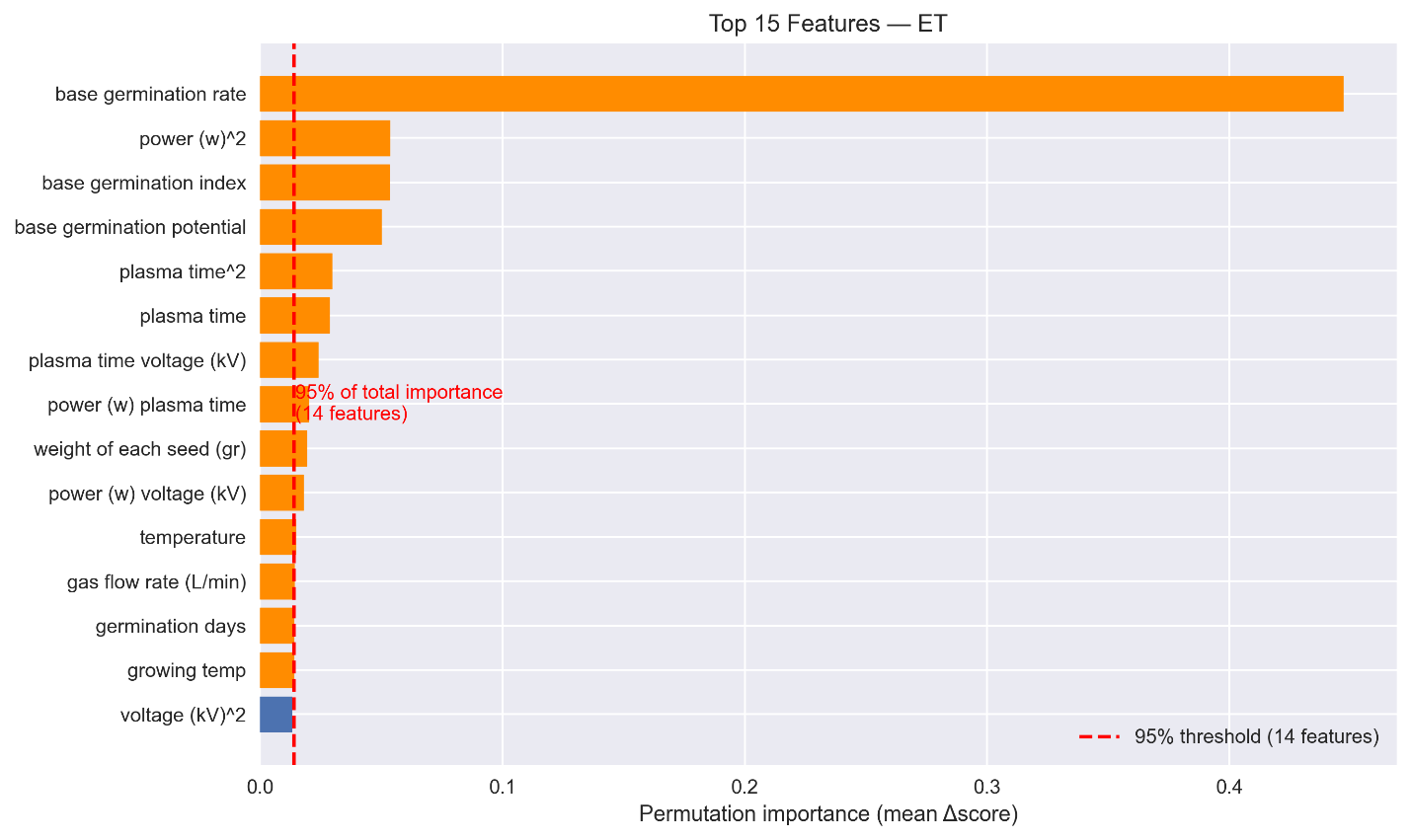}
\caption{Feature selection using permutation importance (ET model).}
\label{fig:figure6}
\end{figure}

Results showed a highly concentrated profile with baseline germination rate emerging as the dominant driver of predictions. Secondary but notable contributions came from baseline germination index, plasma treatment parameters (exposure time, power, voltage), and environmental variables such as water availability and growing temperature. This pattern indicates that ET relies strongly on intrinsic seed vigor while incorporating complementary plasma and environmental factors that modulate treatment response. These findings confirm that CP-induced uplift is driven by both biological traits and controllable plasma parameters \citep{sayahi2024b}.

Retraining ET using this reduced subset showed improved predictive accuracy (table 6). On the training set, the model achieved RMSE = 1.43, MAE = 0.91, and R² = 0.984, which closely matches the full-feature baseline (supplementary figure 8). On the test set, the simplified model slightly improved generalization. RMSE decreased from 3.21 to 3.11, MAE from 2.62 to 2.54, and R² increased from 0.919 to 0.925. Supplementary figure 9 illustrates the actual vs predicted germination uplift plot of this feature-reduced ET model.

\begin{table}[h]
\centering
\caption{Performance of Extra Trees (ET) model trained on top-ranked features (95\% cumulative importance). Metrics are reported for train and test splits.}
\label{tab:table6}
\begin{tabular}{lcccc} 
\toprule
Model & Split & RMSE & MAE & R² \\  
\midrule
ET & Train & 1.43 & 0.91 & 0.984 \\
ET & Test  & 3.11 & 2.54 & 0.925 \\
\bottomrule
\end{tabular}
\end{table}

These results indicate that discarding redundant predictors not only reduces model complexity but also mitigates mild overfitting. The reduced ET model was therefore adopted as the main predictive tool.

\subsection{Species and cultivar-specific performance of ET}

At the seed level (figure 7a), prediction errors varied. Sunflower seeds had the highest MAE (3.80), indicating greater difficulty in capturing their germination dynamics, while radish and soybean had much lower MAEs (1.46 and 2.05, respectively), showing more consistent predictions. Barley (2.58) and tomato (2.33) were intermediate. This suggests that seed-specific physiological traits, such as the variable germination responses in sunflower, as also reported by \cite{sayahi2024b} and \cite{tamosiune2020}, contribute to higher prediction errors while more stable responses in radish lead to more reliable modeling.

\begin{figure}[h]
\centering
\includegraphics[width=0.7\textwidth]{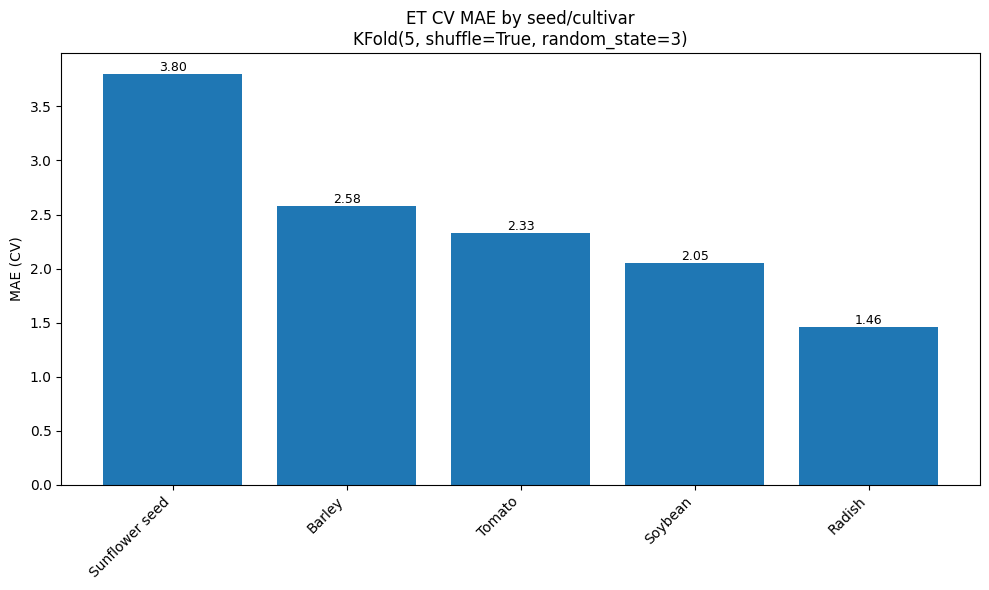}

\vspace{1em} 
\includegraphics[width=0.7\textwidth]{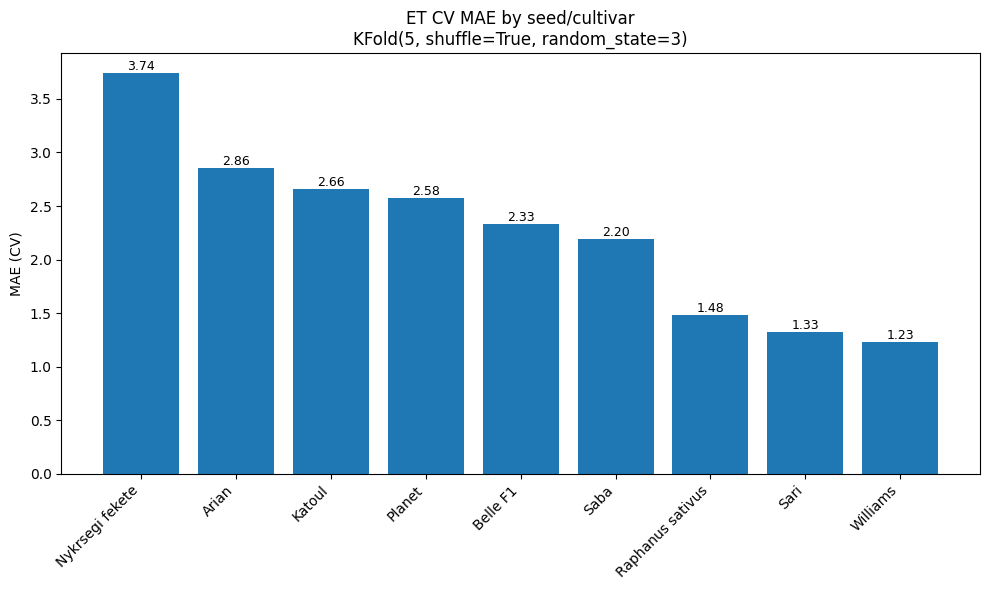}

\caption{Cross-validated MAE of the ET model across different a) seed and b) cultivar types.}
\label{fig:figure7}
\end{figure}

At the cultivar level (figure 7b), a finer-grained picture emerged. Within soybean, the model captured substantial heterogeneity, with cultivars such as Williams (MAE = 1.23) and Sari (1.33) predicted with high precision, while Arian (2.86) showed higher errors. Similarly, among tomato cultivars, Belle F1 (2.33) was modeled more reliably than Katoul (2.66).

Predictability, while broadly high across models, exhibited seed-type dependence, radish emerging as the most tractable and sunflower the most challenging, whereas cultivar-specific physiological and phenotypic heterogeneity further modulated model accuracy, with uniform cultivars yielding consistent fits and variable ones inflating prediction error. From a biological perspective, this variability reflects the interplay between cold plasma treatment effects and the inherent physiology of each cultivar \citep{ling2014}. From a modeling perspective, it highlights the importance of considering cultivar-level differences, as aggregating all seeds can obscure these critical variations, which are important for practical seed technology applications \citep{medeiros2020, sayahi2024b}.

\subsection{Performance of ET under Leave-One-Cultivar-Out (LOCO) validation}

To evaluate the ET beyond cultivars seen during training, we implemented a LOCO cross-validation scheme. This design ensures that in each fold all samples from one cultivar were withheld, simulating a deployment scenario where the model encounters an entirely new genetic background \citep{meredig2018}.

The results (table 7, supplementary figure 10) revealed a substantial decline in predictive accuracy compared with the standard K-Fold CV. Overall LOCO performance dropped to RMSE = 11.62, MAE = 9.28, and R² = --0.06. At the cultivar level, error magnitudes varied markedly. Nyírségi fekete, Planet, and Arian recorded the highest errors (MAE $>$ 12, strongly negative R²), indicating that the model was unable to capture their cold plasma response without prior examples. By comparison, Sari (MAE = 1.19, R² = 0.47) and Belle F1 (MAE = 4.91, R² = 0.49) were predicted with greater reliability. Interestingly, Raphanus sativus and Williams showed negative R² values despite moderate MAE, suggesting that the model had difficulty capturing their specific characteristics.

These findings demonstrate that while ET provides excellent within-cultivar predictions, its generalization to unseen cultivars is limited. This highlights the strong influence of cultivar-specific physiology on plasma-induced germination, consistent with prior work showing that genotype-dependent antioxidant capacity and seed coat morphology shape plasma sensitivity \citep{mohajer2024b, sayahi2024b}. From a modeling perspective, it underscores the need for cultivar-aware features or additional biochemical descriptors to improve cross-cultivar prediction, as relying solely on baseline germination traits and plasma parameters does not capture the genetic determinants of response.

\begin{table}[h]
\centering
\caption{Overall and per-cultivar performance of the Extra Trees (ET) model under Leave-One-Cultivar-Out (LOGO) validation. Values represent mean absolute error (MAE), root mean squared error (RMSE), and coefficient of determination (R²) for out-of-fold predictions.}
\label{tab:table7}
\begin{tabular}{lccc}
\toprule
Cultivar & MAE & RMSE & R² \\
\midrule
Overall LOGO & 9.28 & 11.62 & $-0.06$ \\
\midrule
Nyírségi fekete & 14.11 & 15.99 & $-2.06$ \\
Planet & 12.25 & 13.41 & $-2.77$ \\
Arian & 12.17 & 13.48 & $-0.81$ \\
Raphanus sativus & 9.70 & 11.09 & $-4.04$ \\
Katoul & 6.87 & 9.31 & $-1.00$ \\
Belle F1 & 4.91 & 6.11 & 0.49 \\
Williams & 3.69 & 3.85 & $-4.11$ \\
Saba & 3.32 & 3.51 & $-0.71$ \\
Sari & 1.19 & 1.41 & 0.47 \\
\bottomrule
\end{tabular}
\end{table}

\subsection{Performance on external validation dataset}

Since the feature-reduced ET model achieved robust internal performance, we further validated it on an independent dataset to test its generalizability. The test was done on radish, soybean, barley, and sunflower seeds.

On the external dataset, the ET model achieved an RMSE of 5.68, MAE of 4.67, and R² of 0.524 (table 8). While these scores are weaker compared to internal performance, they still confirm that the model captured a substantial fraction of the variance in uplifted germination across unseen seed--plasma combinations. The residual distribution (figure 8) reveals a tendency to overpredict at lower uplift values and underpredict at higher ones, indicating systematic bias that could be corrected in future iterations with more diverse training data.

\begin{table}[h]
\centering
\caption{External validation performance of the Extra Trees (ET) model on an independent dataset. The table reports root mean squared error (RMSE), mean absolute error (MAE), and coefficient of determination (R²).}
\label{tab:table8}
\begin{tabular}{ccc}
\toprule
RMSE & MAE & R² \\
\midrule
5.68 & 4.67 & 0.524 \\
\bottomrule
\end{tabular}
\end{table}

\begin{figure}[h]
\centering
\includegraphics[width=0.7\textwidth]{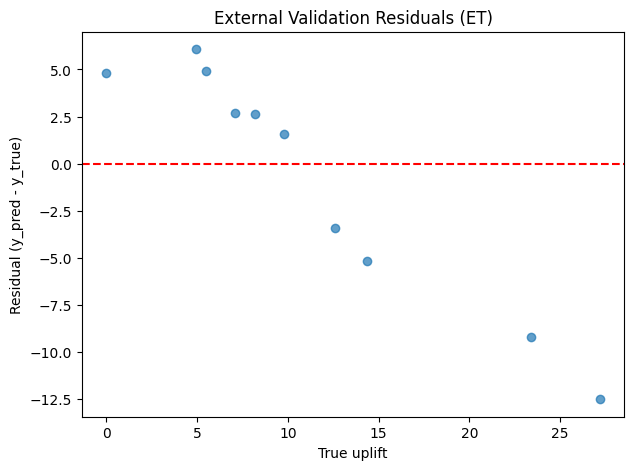}
\caption{Residual plot for external validation of the ET model. Residuals (prediction -- observation) are plotted against true uplift values. The pattern indicates overprediction at lower uplift values and underprediction at higher ones.}
\label{fig:figure8}
\end{figure}

Rather than reflecting model failures, these weaknesses underscore key engineering challenges. Progress will require integrating cultivar-aware descriptors such as seed coat imaging or biochemical markers to better capture intrinsic variability, alongside developing larger, multi-environment CP datasets for training scalable predictive tools. Overcoming these challenges is essential to advance CP seed priming from controlled experimental contexts toward reliable, field-ready commercial deployment \citep{shelar2022}.

\subsection{Biological and engineering drivers of CP-induced uplift}

As shown in figure 9a, the effectiveness of cold plasma priming was strongly dependent on the inherent physiological quality of the seeds. Seeds with lower to intermediate baseline germination rates showed the greatest improvement, benefiting from gains of 20--30\%. In contrast, high-vigor seeds ($>$80\% baseline germination) exhibited minimal improvement and, in some cases, slight reductions. This suggests that cold plasma primarily acts as a vigor amplifier, enhancing the latent potential of seeds rather than improving those that are already high-performing \citep{karimi2024, shahabi2025}.

\begin{figure}[h]
\centering
\includegraphics[width=\textwidth]{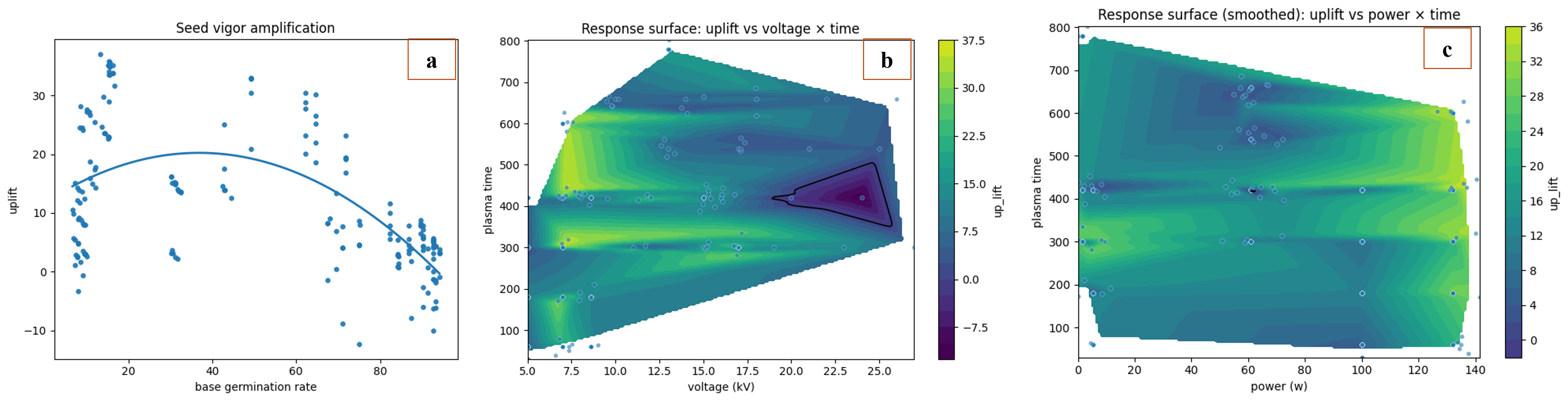}
\caption{a) Relationship between baseline seed vigor and CP-induced germination uplift, b) response surface of germination uplift across voltage--time regimes, c) partial dependence of uplift on power and plasma time.}
\label{fig:figure9}
\end{figure}

The raw voltage--time response surface (figure 9b) provided direct evidence of a safe operating window. Uplift was negligible at low voltages (<7 kV) and short exposures, maximized under moderate conditions (7--15 kV, 200--500s), and reduced under prolonged or high-voltage regimes ($>$20 kV). This pattern is consistent with hormesis, where mild stress stimulates germination but excessive stress causes damage \citep{benabderrahim2024, sayahi2024a}. Complementing this empirical view, model-averaged partial dependence analysis of power and time (figure 9c) revealed that discharge power is the dominant lever. Predicted uplift rose steadily from $\sim$9\% at <40 W to $\sim$16--18\% once power exceeded 100 W, as indicated by the clear horizontal gradient along the power axis. In contrast, plasma time produced only minor shifts, with contour lines running nearly vertical, meaning prolonged exposures alone did not substantially increase uplift. Biologically, this reflects that once seeds receive sufficient reactive species and UV flux per unit time, additional exposure does not enhance germination and may instead trigger oxidative stress, membrane damage, or seed coat cracking \citep{petkova2021, sayahi2024b}. This indicates that efficient priming can be achieved by tuning discharge power rather than extending exposure durations, which adds energy costs and risks of over-treatment without proportional benefits.

SHAP analysis (figure 10) confirmed that cold plasma priming amplifies inherent seed vigor. Low-vigor lots consistently led to positive uplift predictions, while high-vigor lots showed neutral or negative effects, indicating limited potential for further improvement. Among the engineering factors, discharge power was the dominant contributor, reinforcing its role as the key lever for optimizing treatments, similar to the findings reported by \cite{benabderrahim2024} and \cite{perner2024}. Secondary influences included germination environment (temperature) and seed traits (weight), which modulated the effects of vigor and power but did not override them. Plasma time had a marginal effect, consistent with earlier findings that extended exposures offer little additional benefit once effective doses are reached \citep{sayahi2024b, shilpa2024}.

\begin{figure}[h]
\centering
\includegraphics[width=0.7\textwidth]{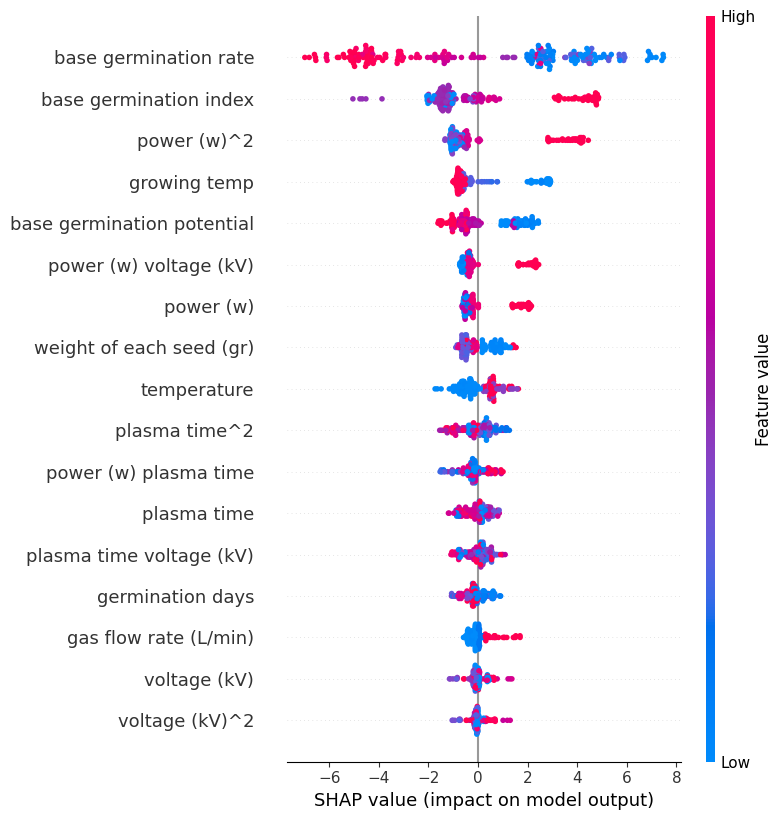}
\caption{Feature contributions to CP-induced uplift integrating biological and engineering factors.}
\label{fig:figure10}
\end{figure}

\subsection{Automated tracking system of ET model}

For practical integration into seed priming systems, machine learning requires reproducibility, traceability, and deployment readiness \citep{macovei2025, shelar2022}. To meet these needs, the reduced-feature ET model was embedded into an MLflow-based automation pipeline, ensuring both computational efficiency and high predictive power.

The MLflow framework automatically tracked all evaluation schemes, including K-Fold cross-validation, Leave-One-Cultivar-Out (LOCO) validation, and external dataset testing, logging hyperparameters, metrics (R², RMSE, MAE), and model artifacts (Supplementary figure 11). Additional MLflow dashboard views, such as experiment tables and run histories are shown in supplementary figure 12. The automated pipeline facilitates reproducible experiment management, real-time monitoring, and deployment-ready optimization.

\section{Conclusion}

This study present a practical, cultivar-aware decision-support framework that predicts CP-induced germination uplift by integrating seed vigor traits with DBD discharge levers (voltage, power, time). ET achieved strong performance (test R² $\approx$ 0.92; RMSE $\approx$ 3.2; MAE $\approx$ 2.6) and improved slightly with a reduced, deployment-ready feature set (R² $\approx$ 0.925), revealing a clear biology--engineering coupling where baseline seed vigor primarily determines responsiveness, while plasma voltage, power, and exposure time act as nonlinear modulators. Radish and soybean remained consistently predictable, whereas sunflower showed greater variability. This compact model is well-suited for embedding in automated priming workflows to shorten trial-and-error and target safe operating windows. External and LOGO performance were moderate, indicating restricted cross-genotype generalization from current, small multi-study datasets. Engineer genotype-aware inputs (e.g., coat imaging/biochemical markers) and curate larger, multi-environment CP datasets to enable robust, scalable optimization and closed-loop treatment design.

\section{CRediT Author Contribution Statement}

\textbf{Saklain Niam}: Conceptualization, Methodology, Investigation, Data Curation, Formal Analysis, Software, Writing -- Original Draft, Visualization. \textbf{Tashfiqur Rahman:} Writing -- Review \& Editing, Data Curation. \textbf{Md. Amjad Patwary, Md. Yasin:} Conceptualization, Resources, Supervision, Writing -- Original Draft, Writing -- Review \& Editing. \textbf{Md. Abu Rayhan, Mukarram Hussain}: Writing -- Review \& Editing. \textbf{Iftekhar Ahmad}: Conceptualization, Supervision, Writing -- review \& editing.

\section{Declaration of Competing Interest}

The authors declare that they have no known competing financial interests or personal relationships that could have appeared to influence the work reported in this paper.

\section{Data Availability}

The datasets, and machine learning code used in this study are available on GitHub at: \url{https://github.com/saqlineniam/mlflow-experimental-tracker-
for-seed-germination-prediction
-from-cold-plasma-priming-parameters}. Additional supporting data are provided as Supplementary Material.

\bibliographystyle{apalike}
\bibliography{references}


\end{document}